\documentclass{article}
\usepackage{spconf,amsmath,graphicx}
\usepackage{amsmath,graphicx}
\usepackage{amsthm}
\usepackage{cite}
\usepackage{graphicx}
\usepackage{amssymb}
\usepackage{amsfonts}
\usepackage{multirow}
\usepackage{mathrsfs}
\usepackage{color}
\usepackage{algorithm}
\usepackage{algorithmic}
\usepackage{multirow}
\usepackage{amsmath}
\usepackage{graphics}
\usepackage{graphicx}
\usepackage{epsfig}
\usepackage{cases}
\usepackage{amsmath,bm}
\usepackage{wasysym}
\usepackage{graphicx}
\usepackage{subfigure}
\usepackage{hyperref}
\hypersetup{ hidelinks = true, }

\newtheorem{mytheorem}{Theorem}

\newtheorem{myproperty}{Property}
\title{Convergence Analysis of Belief Propagation \\for Pairwise Linear Gaussian Models}

\name{Jian Du$^{\dagger}$
Shaodan Ma$^{\star}$,  Yik-Chung Wu$^{\ddag}$,
 Soummya Kar$^{\dagger}$ and  Jos{\'e} M. F. Moura$^{\dagger}$ }
\address{Electrical and Computer Engineering, Carnegie Mellon University$^{\dagger}$, Pittsburgh, PA \\
    Electrical and Computer Engineering, University of Macau$^{\star}$, Macau\\
    Electrical and Electronic Engineering, The University of Hong Kong$^{\ddag}$, Hong Kong}
\begin{document}
\newcommand*{\QEDA}{\hfill\ensuremath{\blacksquare}}
\def\N{{\mathcal{N}}}
\def\B{{\mathcal{B}}}
\def\I{{\textbf{I}}}
\def\diag{{\textrm{diag}}}
\def\i {{ -i}}

%\ninept
%
\maketitle
\begin{abstract}
Gaussian belief propagation (BP) has been widely used for distributed
inference in large-scale networks such as the
smart grid,  sensor networks, and social networks, where local measurements/observations are scattered over a wide geographical area.
One particular case is when two neighboring agents share a common observation.
For example, to estimate voltage in the direct current (DC) power flow model,  the current measurement over a power line is proportional to the voltage difference between two neighboring buses.
When applying the Gaussian BP algorithm to this type of problem, the convergence condition remains an open issue.
In this paper, we analyze the convergence properties of Gaussian BP for this pairwise linear Gaussian model.
We show analytically that the updating information matrix converges at a geometric rate to a unique positive definite matrix with arbitrary positive semidefinite initial value and further provide the necessary and sufficient convergence condition for the  belief mean vector to the optimal estimate.
\end{abstract}

\section{Introduction}\label{Section 1}
Gaussian belief propagation (BP)
 provides an efficiently distributed way to compute the marginal distribution from the joint distribution of  unknown random variables, and it has been adopted in a variety of areas
 such as distributed power state estimation \cite{A1} in power networks, synchronization  \cite{JianClock,cfo,du2013fully} in wireless communication networks \cite{ZhouJSAC,ZhouMag},  cooperative localization in distributed networks \cite{wymeersch2009cooperative}, factor analyzer network \cite{A6}, sparse Bayesian learning \cite{A7}, % inter-cell interference mitigation \cite{A8},
 and peer-to-peer rating in social networks \cite{A9}.
In one particular model of interested  studied in \cite{JianClock, cfo, wymeersch2009cooperative, A9, du2014distributed, du2013fully}),
 two neighboring
agents share a common observation.
In this paper, we name this type of model  pairwise linear Gaussian models.

Although with great empirical success, the  major challenge that hinders Gaussian BP to realize its full potential is the lack of theoretical guarantees of convergence in loopy networks.
Sufficient convergence conditions for Gaussian BP have been developed in \cite{DiagnalDominant,WalkSum1,minsum09}  when the underlying Gaussian distribution is expressed in terms of pairwise connections between scalar variables (also known as Markov random field (MRF)).
However, as demonstrated in \cite{journalversion} the iterative equations for Gaussian BP on MRFs are different from that for distributed estimation problems such as in
 \cite{A1,A2,JianClock,cfo,du2013distributed,du2014distributed},  where  linear measurements  are involved.
Therefore, the existing conditions and analysis methods in
\cite{DiagnalDominant,WalkSum1,minsum09}
  are not applicable to distributed estimation problems.
Though \cite{journalversion} gives the necessary and sufficient condition of  BP for the Gaussian linear model, the type of observation allowed in \cite{journalversion} is not the most general in the sense that it does not allow two neighboring agents to share a common observation.
In this paper, we focus particularly on the convergence analysis of  BP for this pairwise linear Gaussian model.
We
show analytically that the updating of the information matrix converges at a geometric rate to a unique positive definite matrix with arbitrary positive semidefinite initial value and further provide the necessary and sufficient convergence condition for the updating belief mean vector to the optimal estimate.

Note that, in the setup of deterministic unknown parameter  estimation, the distributed algorithm based on the consensus$+$innovations philosophy   proposed in \cite{Kar-SPS, Kar-SIAM} (see also the related family of diffusion algorithms~\cite{Sayed10Diffusion}) converges to the optimal centralized estimator under the assumption of global observability
of the (aggregate) sensing model and connectivity of the inter-agent communication network.
In particular, these algorithms allow
1) the communication or message exchange network to be different from the physical coupling network, and
2)  the  communication network to have arbitrary network structure with cycles (as long as it is connected).
The results in \cite{Kar-SPS, Kar-SIAM} imply that the unknown variables $\textbf x$  can be reconstructed completely at each agent in the network.
For large-scale networks with high dimensional $\textbf x$, it may be impractical to reconstruct $\textbf x$ at every agent.
In \cite[section 3.4]{Kar-thesis}, the author developed approaches to address this problem, where each agent can reconstruct a set of unknown variables that should be larger than the set of variables that influence its local measurement.
This paper studies a different distributed estimation problem when each agent estimates only its own unknown variables under pairwise independence condition of the unknown variables; this leads to lower dimensional data exchanges between neighbors.

\section{Computation Model}\label{hybrid}
Consider a general connected network
of $M$  agents, with $\mathcal{V}=\{1,\ldots, M\}$ denoting the set of agents,
and $\mathcal{E}_{\textrm{Net}} \subset \mathcal{V} \times  \mathcal{V}$ as the set of all undirect communication links in the network, i.e., if $i$ and $j$ are within the communication range, $(i, j) \in \mathcal{E}_{\textrm{Net}}$.
The  local observations, $\textbf{y}_{i,j}$, between agents $i$ and $j$ are modeled by a pairwise Gaussian linear model:
\begin{equation} \label{linear}
\textbf{y}_{i,j} =
\textbf{A}_{j,i}\textbf{x}_i
+\textbf{A}_{i,j}\textbf{x}_j
+ \textbf{z}_{i,j},
\end{equation}
%$\textbf{y}_n$ is the local observation at node $n$,
where
$\textbf{A}_{j,i}$ and $\textbf{A}_{i,j}$ are the known coefficient matrices with full column rank,
$\textbf{x}_i$ and $\textbf{x}_{j}$ are the local unknown vector parameters at agent $i$ and $j$ with  dimension $N_i \times 1$ and $N_j \times 1$, and with the prior distribution $p(\textbf{x}_i)\sim \mathcal{N}(\textbf{x}_i|\textbf{0},\textbf{W}_{i})$ and
$p(\textbf{x}_j)\sim \mathcal{N}(\textbf{x}_j|\textbf{0},\textbf{W}_{j})$
and $\textbf{z}_{i,j}$ is the additive noise with distribution $\textbf{z}_{i,j}\sim \mathcal{N}(\textbf{z}_{i,j}|\textbf{0},\textbf{R}_{i,j})$.
It is assumed that
$p(\textbf{x}_i, \textbf{x}_j)=p(\textbf{x}_i)p(\textbf{x}_j)$
and
$p(\textbf{z}_{i,j},\textbf{z}_{s,t})
=p(\textbf{z}_{i,j})p(\textbf{z}_{s,t})$ for $\{i,j\}\neq \{s,t\}$.
The goal is to estimate $\textbf{x}_i$, based on $\textbf{y}_{i,j}$, $p(\textbf{x}_i)$ and $p(\textbf{z}_{i,j})$ for all $\textbf x_i\in
\mathcal V$.
Note that in (\ref{linear}), $\textbf{y}_{i,j}=\textbf{y}_{j,i}$.

In centralized estimation, all the observations $\textbf{y}_{i,j}$ at different agents are forwarded to a central processing unit.
Define  vectors $\textbf{y}$, $\textbf x$ and $\textbf z$ as the stacking of $\textbf{y}_{i,j}$,  $\textbf x_{i}$ and $\textbf{z}_{i,j}$ in ascending order first with respect to $i$ and then on $j$, respectively;  then we obtain
%\begin{equation}\label{blocklinear}
$\textbf{y}= \textbf{A}\textbf x + \textbf z,$
%\end{equation}
where
$\textbf{A}$ is constructed from $\textbf{A}_{n,i}$, with specific arrangement depending on the network topology.
Assuming $\textbf{A}$ is a full column rank matrix, and since
$\textbf z$ is a  Gaussian random vector, the optimal estimate $\hat{\textbf x}\triangleq[\hat{\textbf x}_1^T,\ldots,\hat{\textbf x}_M^T]^T$ of $\textbf x$   is given by
\begin{eqnarray}\label{Central}
\hat{\textbf x}
=
(\textbf{W}^{-1}+ \textbf{A}^T\textbf{R}^{-1}\textbf{A})^{-1}\textbf{A}^T\textbf{R}^{-1} \textbf{y},
\end{eqnarray}
where $\textbf{W}$ and $\textbf{R}$ are block diagonal matrices containing $\textbf{W}_{i,j}$ and $\textbf{R}_{i,j}$ as their diagonal blocks, respectively.
Although well-established, the drawbacks of the centralized estimation in large-scale networks include 1) the transmission of $\textbf{y}_{i,j}$, $\textbf{A}_{i,j}$ and $\textbf R_{i,j}$ from
 peripheral agents to the computation center imposes huge communication overhead;
2) knowledge of the global network topology is  needed in order to construct $\textbf{A}$; and
3) the computation burden at the computation center scales up with the cubic of the dimension of the matrix inverse in (\ref{Central})  with   complexity order $\mathcal{O}((\sum_{i=1}^{|\mathcal{V}|}N_i)^3)$.

The joint distribution  $p\left(\textbf x\right)p\left(\textbf{y}|\textbf x\right)$ is first written as the product of the  prior distribution and the likelihood function as
%\begin{equation}\label{jointpost}
$$p\left(\textbf x\right)p\left(\textbf{y}|\textbf x\right) =
\prod_{i\in \mathcal{V}}
\underbrace{p\left(\textbf x_i\right)}_{\triangleq f_{i}}
%\underbrace{p(\textbf x_n) }
%_{\triangleq g_n}
\prod_{i\in \mathcal{V}}
\underbrace{p(\textbf{y}_{i,j}| \textbf{x}_i, \textbf{x}_j, \{i,j\}\in  \mathcal{E}_{\textrm{Net}})}_{\triangleq f_{i,j}}.$$
%\end{equation}
To facilitate the derivation of the distributed inference algorithm,
the factorization above
is expressed in terms of a
factor graph,
where  every variable vector $\textbf{x}_i$ is represented by a  variable node
and the probability distribution of a vector variable or a group of vector variables is represented by a factor node.
A variable node is connected to a factor node if the variable is involved in that particular factor.
It
involves two types of messages:
One is the message
from a factor  node with function $f$ to its neighboring variable node $\textbf x_i$, defined as
\begin{equation}\label{BPf2v1}
m^{(\ell)}_{{f} \to i}(\textbf{x}_i)
= \!\! \int\!\!\! \cdots \!\!\!\int
\!\! f \times\!\!\!\!\!\!\!
\prod_{j\in\B( f)\setminus i}
\!\!\!\!\!\!
m^{ (\ell)}_{j \to  f}(\textbf x_j)
\,\mathrm{d}\{\textbf x_n\}_{n\in\B(f)\setminus i},
\end{equation}
where $\B(f)$ denotes the set of neighboring variable nodes of factor node $f$ on the factor graph.
The other type of message is from factor node $\tilde{f}$, which denotes a likelihood function or prior distribution, to its neighboring variable node $\textbf x_i$
and it is defined as
\begin{equation} \label{BPv2f1}
m^{(\ell)}_{j \to f}(\textbf x_i)
=
\prod_{\tilde{f}\in \B(j)\setminus f}m^{(\ell-1)}_{\tilde{f}\to j}(\textbf x_j),
\end{equation}
where $\B(j)$ denotes the set of neighbouring factor nodes  of $\textbf x_j$,
and $m^{(\ell-1)}_{\tilde f\to j}(\textbf x_j)$ is the   message  from $\tilde f$ to $\textbf x_j$ at time $l-1$.
The process iterates between equations (\ref{BPv2f1}) and (\ref{BPf2v1}).
At each iteration $\ell$, the approximate marginal distribution, also named belief, on $\textbf{x}_i$ is computed locally at $\textbf{x}_i$ as
\begin{equation} \label{BPbelief}
b_{\textrm{BP}}^{(\ell)}(\textbf{x}_i)
 = \prod_{ f\in \B(i)} m^{(\ell)}_{ f \to i}(\textbf{x}_i).
\end{equation}
%\begin{figure}[t]\label{SystemModel}
%  \centering
%\mbox{\subfigure[]{\epsfig{figure=./Fig1,width=2in}}\label{Network} }\\
%\mbox{ \subfigure[]{\epsfig{figure=./Fig2,width=2.8in}}\label{MRF} }\\
%\caption{(a) A physical network with $4$ nodes, where $\textbf{x}_i$ is the local unknown vector and
%two variables are linked by an edge if there is a local pairwise observation
%that follows (\ref{linear});
%(b) The corresponding factor graph of Fig.~1 (a) with $ \psi_{n} \left(\textbf x_n, \left\{\textbf{y}_i\right\}_{i\in n\cup\mathcal{I}(n) }\right)$ and $\psi_{n,i} \left(\textbf x_n, \textbf{x}_i\right)$ defined in
%(\ref{MRF-local}) and (\ref{MRF-pair}), respectively.}
%\end{figure}
%\begin{figure}[t]
%  \centering
%{\epsfig{figure=./Figures/message_evolution_figure,width=6.6in}}
%\caption{Messages computation and transmission at the $l^{\textrm{th}}$ iteration.}
%\label{Iteration}
%\end{figure}
%Following the computation rule of  (\ref{BPf2v1}),
%it can be easily shown  that
%message from  a factor that denotes prior information is a constant, i.e.,
%%\begin{equation}
%$m^{(\ell)}_{{f}_i \to i}(\textbf{x}_i)
%= p(\textbf x_i),\quad \forall \ell=0,1,\ldots\nonumber.$
%%\end{equation}
%Besides, from (\ref{BPf2v1}) and (\ref{BPv2f1}), message $m^{(\ell)}_{i \to f_i}(\textbf{x}_i)$ will never be used for message updating.
%Therefore, we next will focus on message exchange between factor denoting likelihood function and its
%neighboring variables.
It can be  shown  that the message  from factor node $f_{i,j}$ to variable node $i$ is given by \cite{journalversion}
\begin{equation} \label{f2v}
m^{(\ell)}_{f_{i,j} \to i}(\textbf{x}_i)\propto
\exp
\big\{-\frac{1}{2}
||\textbf{x}_i- \textbf{v}^{(\ell)}_{f_{i,j}\to i}||^2
_{\textbf{C}^{(\ell)}_{f_{i,j}\to i}}
\big\},
\end{equation}
where $\textbf{C}_{f_{i,j}\to j}^{(\ell-1)}$ and $ \textbf{v}_{f_{i,j}\to j}^{(\ell-1)}$ are the message covariance matrix and mean vector  received at variable node $j$ at the $l-1$ iteration
with
\begin{equation}\label{Cov}
\begin{split}
\left[\textbf{C}^{(\ell)}_{f_{i,j}\to i} \right]^{-1}
=
\textbf{A}_{j,i}^T
\left[ \textbf{R}_{i,j}
+ \textbf{A}_{i,j}\textbf{C}^{(\ell)}_{j\to f_{i,j}}\textbf{A}_{i,j}^T \right]^{-1}
\textbf{A}_{j,i}.
\end{split}
\end{equation}
and
\begin{equation}\label{f2vmm}
\begin{split}
\textbf{v}^{(\ell)}_{f_{i,j}\to i}
\!=&
\textbf{A}_{j,i}^T
\left[ \textbf{R}_{i,j}\!
+ \textbf{A}_{i,j}\textbf{C}^{(\ell)}_{j\to f_{i,j}}\textbf{A}_{i,j}^T \right]^{-1}
\!\!\left(\textbf{y}_{i,j}\!\!- \textbf{A}_{i,j}
\textbf{v}^{(\ell)}_{j\to f_{i,j}}\!\right).
\end{split}
\end{equation}
Furthermore, the general expression for the message from variable node $j$ to factor node $f_{i,j}$ is
\begin{equation} \label{BPvs2f1}
m^{(\ell)}_{j \to f_{i,j}}(\textbf x_j) \propto
\exp
\big\{-\frac{1}{2}
||\textbf x_j- \textbf{v}^{(\ell)}_{j\to f_{i,j}}||^2
_{\textbf{C}^{(\ell)}_{j\to f_{i,j}}}
\big\},
\end{equation}
where $\textbf{C}_{j\to f_{i,j}}^{(\ell)}$ and $ \textbf{v}_{j\to f_{i,j}}^{(\ell)}$ are the message covariance matrix and mean vector  received at variable node $j$ at the $\ell$-$\textrm{th}$ iteration, with the information matrix computed as
\begin{equation} \label{v2fV}
\big[\textbf{C}^{(\ell)}_{j \to f_{i,j}}\big]^{-1}
= \textbf{W}_j^{-1} +
\sum_{f_{k,j}\in\B(j)\setminus f_{i,j}}
\big[\textbf{C}_{f_{k,j}\to j}^{(\ell-1)}\big]^{-1}.
\end{equation}
and the mean vector is
\begin{equation}\label{v2fm}
\textbf{v}^{(\ell)}_{j\to f_{i,j}}=
\textbf{C}^{(\ell)}_{j\to f_{i,j}}
\bigg[\!\!\!
\sum_{f_{k,j}\in\B(j)\setminus f_{i,j}}\!\!\!\!\!
\big[\textbf{C}_{f_{k,j}\to j}^{(\ell-1)}\big]^{-1}
\textbf{v}^{(\ell-1)}_{f_{k,j}\to j}\bigg],
\end{equation}

Following Lemma 2 in \cite{journalversion}, we know  that   setting the initial information matrix  $[\textbf{C}_{f_{k,j}\to i}^{(0)}]^{-1}\succeq \textbf{0}$  for all $k\in \mathcal{V}$ and $j\in \mathcal B(k)$
guarantees  $[\textbf{C}^{(\ell)}_{j\to f_{i,j}}]^{-1}\succ \textbf{0}$ for $l \geq 1$.
Therefore,
let the initial messages at factor node $f_{k,j}$ be in Gaussian function forms with  covariance $[\textbf{C}_{f_{k,j}\to j}^{(0)}]^{-1} \succeq \textbf{0}$ for all $k \in \mathcal{V}$ and $j \in \mathcal{B}(f_{k,j})$.
Then
all the messages $m^{(\ell)}_{j \to f_{i,j}}(\textbf x_j)$ and
$m^{(\ell)}_{f_{i,j} \to i}(\textbf{x}_i)$ exist and are in Gaussian form.
Furthermore, during each round of message passing, each agent can compute the belief for $\textbf x_i$
using (\ref{BPbelief}), which can be easily shown to be
\begin{equation}
b_{i}^{(l)}(\textbf x_i)\sim  \mathcal{N}(\textbf x_i|\bm \mu_i^{(l)}, \textbf P_{i}^{(l)}),
\end{equation}
with the
inverse of the covariance matrix
\begin{equation} \label{beliefP}
 \big[\textbf P_{i}^{(l)}\big]^{-1} = \sum_{f_{i,j}\in\mathcal B(f_{i,j})}\big[\textbf C_{f_{i,j}\to i}^{(l)}\big]^{-1},
\end{equation}
and mean vector
\begin{equation}\label{beliefu}
 \bm \mu_i^{(l)} = \![\!\!\sum_{f_{i,j}\in\mathcal B(f_{i,j})}\!\!\!\!\!\big[\textbf C_{f_{i,j}\to i}^{(l)}\big]^{-1}]^{-1}
 \!\!\!\!\sum_{j\in\mathcal B(f_{i,j})}\!\!\!\!\!\big[\textbf C_{f_{i,j}\to i}^{(l)}\big]^{-1}\textbf v^{(l)}_{f_{i,j}\to i}.
\end{equation}

The iterative algorithm based on BP is summarized as follows.
The algorithm is started by setting the message from factor node to variable node as
$m^{(0)}_{f_{i,j} \to  i}(\textbf x_i)=\mathcal{N}\left(\textbf x_i;\bm v^{(0)}_{f_{i,j}\to i}, \bm C_{f_{i,j}\to i}^{(0)}\right)$
with a random initial vector $\bm v^{(0)}_{f_{i,j}\to i}$  and $\left[\bm C_{f_{i,j}\to i}^{(0)})\right]^{-1}\succeq \bm 0$.
At each round of message exchange, every variable node computes the outgoing messages to factor nodes
according to (\ref{v2fV}) and (\ref{v2fm}).
After receiving the messages from its neighboring variable nodes, each factor node computes its outgoing messages according to (\ref{Cov}) and (\ref{f2vmm}).
Such iteration is terminated when (\ref{beliefu}) converges (e.g., when $\|\bm\mu_i^{(\ell)}-\bm\mu_i^{(\ell-1)}\|<\eta$, where $\eta$ is a threshold) or the maximum number of iterations is reached.
Then the estimate of $\textbf x_i$ of each node is obtained as in  (\ref{beliefu}).

 %In practical networks, there is neither factor nodes nor variable nodes. The two kinds of messages
%$m^{(\ell)}_{i \to f_{i,j}}(\textbf x_i) $ and $m^{(\ell)}_{f_{i,j} \to j}(\textbf x_j)$
%are computed locally at node $i$, and only mean vector $ \textbf v^{(\ell)}_{f_{i,j} \to  j}(\textbf x_j)$
%and covariance matrix
%$\textbf C_{f_{i,j}\to j}^{(\ell)}(\textbf x_j)$
%are passed from node $i$ to node $j$ during each round of message exchange of BP.
%It can be seen the algorithm is fully distributed and each node only needs to exchange limited information with neighboring nodes.

\section{ Convergence Analysis}\label{analysis}
The challenge of deploying the BP algorithm for large-scale networks is determining whether it will converge.
In particular, it
is generally known that, if the factor graph contains cycles, the BP algorithm may
diverge.
Thus, determining  convergence conditions for the BP algorithm is very important.
%Besides, easily achieved sufficient convergence condition is of great importance for arbitrary network topology in practical applications.
Sufficient conditions for the convergence of Gaussian BP with scalar variable in loopy graphs are available in \cite{DiagnalDominant, WalkSum1} for Markov random fields.
Unfortunately, as first pointed out in \cite{journalversion}, the convergence analysis for the Gaussian Markov random field  and for the Gaussian linear model are quite different due to different iteration equations.
Though \cite{journalversion} gives the necessary and sufficient condition of  BP for the Gaussian linear model, the type of observations allowed in \cite{journalversion} (e.g., equation (1) in \cite{journalversion}), is not the most general in the sense that it does not allow two neighboring agents to share a common observation as in equation (1) in this paper.
In the following, we provide the convergence analysis of Gaussian BP for the pairwise linear Gaussian model.

Due to the recursively updating property of  $m_{j\to f_{i,j}}^{(\ell)}(\textbf x_j)$ and $m_{f_{i,j}\to i}^{(\ell)}(\textbf{x}_i)$ in (\ref{BPvs2f1}) and (\ref{f2v}), the message evolution can be simplified by combining these two types of messages into a single one.
By substituting
 $ \big[\textbf{C}^{(\ell)}_{j \to f_n}\big]^{-1}$ in (\ref{v2fV}) into   (\ref{Cov}), the updating of the message covariance matrix inverse, named message information matrix in the following, can be denoted as
 \begin{eqnarray}\label{CovFunc}
[\textbf{C}^{(\ell)}_{f_{i,j}\to i} ]^{-1}
&=&
\textbf{A}_{j,i}^T
\big[ \textbf{R}_{i,j}
+ \textbf{A}_{i,j}
\big[\textbf{W}_j^{-1}\nonumber\\
&+&
\sum_{f_{k,j}\in\B(j)\setminus f_{i,j}}
\big[\textbf{C}_{f_{k,j}\to j}^{(\ell-1)}\big]^{-1}\big]^{-1}
\textbf{A}_{i,j}^T \big]^{-1}
\textbf{A}_{j,i}\nonumber\\
&\triangleq &
\mathcal{F}_{n\to i}
\big(\{
\big[\textbf{C}_{f_{k,j}\to j}^{(\ell-1)}\big]^{-1}\}_{f_{k,j}\in\B(j)\setminus f_{i,j}}
  \big).
\end{eqnarray}
Observing that $\textbf{C}_{f_{i,j}\to i}^{(\ell)}$ in (\ref{CovFunc}) is independent of $\textbf{v}^{(\ell)}_{f_{i,j}\to i} $, the other type of updating information, we  first focus on the convergence property of $[\textbf{C}_{f_n\to i}^{(\ell)}]^{-1}$.

%\begin{figure}[!ht]
%  \centering
%{\epsfig{figure=./Figures/Convariance,width=3.2in}}
%\caption{ Illustration for the construction of (\ref{CovFunc3}) from (\ref{CovFunc}).}
%\label{MSE-SNR}
%\end{figure}
To consider the updates of all message
information matrices, we {{introduce} the following definitions.
Let
%\begin{equation}
${\textbf{C}}^{(\ell-1)}
\triangleq
\texttt{Bdiag}
(\{[\textbf{C}_{f_{i,j}\to i}^{(\ell-1)}]^{-1}\}_{i\in \mathcal{V},\{i,j\}\in  \mathcal{E}_{\textrm{Net}}}\nonumber$
%\end{equation}
 be
a block diagonal  matrix with diagonal blocks
being the   message information matrices in the network at time $l-1$
with index arranged in ascending order first on $i$ and then on $j$.
Using the definition of $\textbf{C}^{(\ell-1)}$, the term $\sum_{f_{k,j}\in\B(j)\setminus f_{i,j}}
\big[\textbf{C}_{f_{k,j}\to j}^{(\ell-1)}\big]^{-1}$ in (\ref{CovFunc}) can be written as $\boldsymbol{\Xi}_{i,j} \textbf{C}^{(\ell-1)} \boldsymbol{\Xi}_{i,j}^T$, where $\boldsymbol{\Xi}_{i,j}$  selects appropriate components from $\textbf{C}^{(\ell-1)}$ to form the summation.
%Further, define
%%$\textbf{H}_{n,i}=[\{ \textbf{A}_{n,j} \}_{j\in \mathcal B(f_n) \backslash i}]$,
%$\boldsymbol{\Psi}_{n,i}  = \texttt{Bdiag} (\!\{\textbf{W}_j ^{-1} \}_{j\in \mathcal B(f_n) \backslash i}\!)$ and $\textbf{K}_{n,i}\!\!=\!\!\texttt{Bdiag} (\!\{ \boldsymbol{\Xi}_{n,j} \}_{j\in \mathcal B(f_n) \backslash i}\!) $, all with component blocks arranged with ascending order on $j$.  Then (\ref{CovFunc}) can be written as
\begin{equation}\label{CovFunc3}
\begin{split}
    \left[\textbf{C}^{(\ell)}_{f_{i,j}\to i} \right]^{-1}
    =&\textbf{A}_{j,i}^T\big \{\textbf{R}_{i,j}+ \textbf{A}_{i,j}[\textbf W_j^{-1} \\
   & + \boldsymbol{\Xi}_{i,j} \textbf{C}^{(\ell-1)} \boldsymbol{\Xi}_{i,j}^T ]^{-1} \textbf{A}_{i,j}^T \big\} ^{-1}\textbf{A}_{j,i}.
\end{split}
\end{equation}
\vspace{-1em}

We define the function $\mathcal{G}\triangleq\{\mathcal{G}_{1\to k}, \ldots, \mathcal{G}_{n\to i}, \ldots,\\ \mathcal{G}_{n \to M}\}$ that updates
${\textbf{C}}^{(\ell)} = \mathcal{G}({\textbf{C}}^{(\ell-1)}) $.
Then, by stacking $\big[\textbf{C}_{f_{i,j}\to i}^{(\ell)}\big]^{-1}$ on the left side of
(\ref{CovFunc3}) for all $n$ and $i$ as the block diagonal matrix $\textbf{C}^{(\ell)}$, we obtain
\begin{eqnarray}\label{CovFunc5}
% \nonumber to remove numbering (before each equation)
  \textbf{C}^{(\ell)}
  &=& \textbf{A}^T \left [\textbf{R}+ \textbf{H}\left(\textbf{W} + \boldsymbol{\Xi} \textbf{C}^{(\ell-1)} \boldsymbol{\Xi}^T \right)^{-1} \textbf{H}^T \right] ^{-1}\textbf{A}, \nonumber\\
   &\triangleq& \mathcal G(\textbf{C}^{(\ell-1)}),
\end{eqnarray}
where $\textbf{A}$, $\textbf{R}$, $\textbf{H}$,
$\textbf{W}$,  and $\boldsymbol{\Xi}$ are block diagonal matrices with block elements $\textbf{A}_{j,i}$, $\textbf R_{i,j}$, $\textbf{A}_{i,j}$, $\textbf{W}_j $, and $\boldsymbol{\Xi}_{i,j}$, respectively, arranged in ascending order, first on $n$ and then on $i$ (i.e., the same order as $[\textbf{C}^{(\ell)}_{f_n \rightarrow i}]^{-1}$ in $\textbf{C}^{(\ell)}$).
We first present  properties of the updating operator $\mathcal{G}(\cdot)$,
where
the  proof  follows that in  \cite{journalversion}.

\begin{myproperty} \label{P_FUN}
 The updating operator $\mathcal{G}(\cdot)$ satisfies the following properties:
\end{myproperty}

\noindent P \ref{P_FUN}.1:
$\mathcal{G}(\textbf{C}^{(\ell)}) \succeq \mathcal{G}(\textbf{C}^{(\ell-1)})$, if $\textbf{C}^{(\ell)} \succeq \textbf{C}^{(\ell-1)}\succeq \textbf{0}$.

\noindent P \ref{P_FUN}.2: $\alpha\mathcal{G}(\textbf{C}^{(\ell)}) \succ  \mathcal{G}(\alpha \textbf{C}^{(\ell)})$
and
$\mathcal{G}(\alpha^{-1}\textbf{C}^{(\ell)}) \succ  \alpha^{-1}\mathcal{G}(\textbf{C}^{(\ell)})$, if $\textbf{C}^{(\ell)} \succ \textbf{0}$ and $\alpha>1$.

\noindent P \ref{P_FUN}.3:
Define
$\textbf{U}\triangleq \textbf{A}^T  \textbf R^{-1}\textbf{A}$
and $\textbf{L}\triangleq
\textbf{A}^T \Big [  \textbf R+ \textbf{H}\textbf W^{-1}\textbf{H}^T \Big] ^{-1}\!\!\textbf{A}$.
With arbitrary $\textbf{C}^{(0)}\succeq \textbf{0}$,
$\mathcal{G}(\textbf{C}^{(\ell)})$ is bounded by
$\textbf{U} \succeq  \mathcal{G}(\textbf{C}^{(\ell)})\succeq \textbf{L}\succ \textbf{0}$ for $l\geq 1$.

In this paper, $\textbf{X} \succeq\textbf{Y}$ ($\textbf{X} \succ \textbf{Y}$) means that $\textbf{X} - \textbf{Y}$ is positive semidefinite (definite).
%\noindent
%\begin{mylemma} \label{fixed}
%(Brouwer's Fixed-Point Theorem \cite{FixedPoint})
% Every continuous mapping $\mathcal{F}$ of a compact, convex,
%and nonempty set has a fixed point.
%\end{mylemma}
Note $\mathcal G$ is different from the function $\mathcal F$ in \cite{ICASSP17}.
However, as demonstrated in \cite{ICASSP17}, if a function
$\mathcal G$ satisfies Property 1,
we can establish the convergence property
for $\textbf C^{(\ell)}$ given by the following
Theorem with detailed provided in \cite{journalversion}.
\begin{mytheorem}\label{RateCov}
With the initial covariance matrix set to be an arbitrary p.s.d. matrix, i.e., $[\textbf{C}^{(0)}_{f_n\to i}]^{-1}\succeq \textbf{0}$,
the sequence $\{\textbf{C}^{(\ell)}\}_{l=0,1,\ldots}$ converges at a double exponential rate to a unique p.d. matrix.
%\\ there is a constant $q>0$ such that
%$||\textbf{C}^{\ast} - \textbf{C}^{(\ell)} ||
%\leq
%q ||\textbf{C}^{\ast}- \textbf{C}^{(\ell-1)} || $ for all $l=0, 1, 2, \ldots $.
%More specially, if $0\leq q<1$, the convergence rate is Q-linear; otherwise if $q>1$, the convergence rate is Q-sublinear.
\end{mytheorem}
%According to  Theorem \ref{RateCov}, the covariance matrix $\textbf{C}^{(\ell)}$ converges    if all initial information matrices are
% p.s.d., i.e.,  $\big[\textbf{C}^{(0)}_{f_{i,j}\to i}\big]^{-1}\succeq \textbf{0}$
% for all $i \in \mathcal V$ and $j\in \mathcal B(i)$.
%{Note that, for the Gaussian Markov random field, the information matrix does not necessarily converge for all initial non-negative value (in the scalar variable case) as shown in \cite{WalkSum1,minsum09}.}
%Moreover, due to the computation of $[\textbf{C}^{(\ell)}_{f_{i,j}\to i}]^{-1}$  being independent of the local observations $\textbf{y}_n$,
%as long as the network topology does not change, the converged value  $[\textbf{C}^{\ast}_{f_{i,j}\to i}]^{-1}$ can be precomputed offline and stored at each node, and there is no need to re-compute $[\textbf{C}^{\ast}_{f_{i,j}\to i}]^{-1}$ even if $\textbf{y}_{i,j}$ varies.
%Theorem \ref{RateCov} also demonstrates that the sequence $\{\textbf{C}^{(\ell)}\}_{l=1,...}$ converges at a geometric rate (the distance between $\textbf{C}^{(\ell)}$ and $\textbf{C}^{\ast}$ decreases exponentially) before $\textbf{C}^{(\ell)}$ enters $\textbf{C}^{\ast}$'s neighborhood, which can be chosen arbitrarily small.

Thus, if we  choose $[\textbf{C}_{f_{i,j}\to j}^{\left(0\right)}]^{-1}\succeq \textbf{0}$
for all $j\in \mathcal V$ and $i\in \mathcal B\left(j\right)$, then
$\left[\textbf{C}_{f_{i,j}\to j}^{\left(\ell\right)}\right]^{-1}$
 converges at a double exponential rate to a unique p.d. matrix
$\left[\textbf{C}_{f_{i,j}\to j}^{\ast}\right]^{-1}$.
Furthermore, according to (\ref{v2fV}),
$
\big[\textbf{C}^{(\ell)}_{j \to f_{i,j}}\big]^{-1}$ also converges to a p.d. matrix once
$\big[\textbf{C}_{f_{k,j}\to j}^{(\ell-1)}\big]^{-1}$ converges; the  converged value is denoted by
$
\big[\textbf{C}^{\ast}_{j \to f_{i,j}}\big]^{-1}$.
Then, for arbitrary initial value
$\textbf{v}^{\left(0\right)}_{f_{k,j}\to j}$, the evolution of
$\textbf{v}^{\left(\ell\right)}_{j\to f_n}$ in
(\ref{v2fm}) can be written in terms of the limit message information matrices as
\begin{equation}\label{v2fm1}
\textbf{v}^{(\ell)}_{j\to f_{i,j}}=
\textbf{C}^{\ast}_{j\to f_{i,j}}
\bigg[
\sum_{f_{k,j}\in\B(j)\setminus f_{i,j}}
\big[\textbf{C}_{f_{k,j}\to j}^{\ast}\big]^{-1}
\textbf{v}^{(\ell-1)}_{f_{k,j}\to j}\bigg].
\end{equation}
Using (\ref{f2vmm}), and
replacing indices $j$, $i$ with $k$, $j$ respectively,
$\textbf{v}^{\left(\ell-1\right)}_{f_{k, j}\to j}$  is given by
\begin{equation}\label{f2vmm2}
\begin{split}
\textbf{v}^{(\ell)}_{f_{k,j}\to j}
=&
\textbf{A}_{k,j}^T
\left[ \textbf{R}_{k,j}
+ \textbf{A}_{j,k}\textbf{C}^{\ast}_{k\to f_{k,j}}\textbf{A}_{j,k}^T \right]^{-1}
\\
&\times
\left(\textbf{y}_{k,j}- \textbf{A}_{j,k}
\textbf{v}^{(\ell)}_{k\to f_{k,j}}\right).
\end{split}
\end{equation}
Putting (\ref{f2vmm2}) into
 (\ref{v2fm1}), we have
\begin{equation}\label{v2fm3}
\textbf{v}^{\left(\ell\right)}_{j\to f_{i,j}}=
\textbf{b}_{j \to f_{i,j}}-
\textbf{C}^{\ast}_{j\to f_{i,j}}\!\!\!\!\!\!\!\!
\sum_{f_{k,j}\in\B\left(j\right)\setminus f_{i,j}}\!\!\!\!\!\!
\textbf{C}^{\ast}_{f_{kj}\to j}
\textbf{M}_{k,j}
\textbf{A}_{j,k}
\textbf{v}^{(\ell)}_{k\to f_{k,j}},
\end{equation}
where $\textbf{b}_{j\to f_{i,j}}=
\textbf{C}^{\ast}_{j\to f_{i,j}}
\sum_{f_{k,j}\in\B\left(j\right)\setminus f_{i,j}}
\textbf{M}_{k,j}
\textbf{y}_k $
and
$\textbf M_{k,j}=\textbf{A}_{k,j}^T\left[ \textbf{R}_{k,j}
+ \textbf{A}_{j,k}\textbf{C}^{\ast}_{k\to f_{k,j}}\textbf{A}_{j,k}^T \right]^{-1}$.
The above equation for all $j\in \N(i)$ cases can be further written in a compact form as
\begin{equation}\label{v2fm4}
\textbf{v}^{\left(\ell\right)}_{j}=
\textbf{b}_{j} -
\textbf{Q}_{j}
\textbf{v}^{\left(\ell-1\right)},
\end{equation}
with the column vector
$\textbf{v}^{(\ell)}_j$ containing all
$\{\textbf{v}^{(\ell)}_{j\to f_{i,j}}\}_{i\in\N(j)}$
as subvectors
with ascending index   on $i$.
Similarly,
$\textbf{b}_{j}$
containing all
$\{\textbf{b}_{j\to f_{i,j}}\}_{i\in\N(j)}$ as subvectors
with ascending index   on  $i$, and
$\textbf{v}^{\left(\ell-1\right)}$
 containing  $\textbf{v}^{\left(\ell-1\right)}_{k\to f_{k,j}}$ for all $f_{k,j}\in\B\left(j\right)\setminus f_{i,j}$ as subvectors
with ascending index first on $z$ and then on $k$.
The matrix   $\textbf{Q}_{j}$
is a block matrix with component blocks $\textbf 0$ and
$\textbf{C}^{\ast}_{j\to f_{i,j}}$ where
$f_{k,j}\in\B\left(j\right)\setminus f_{i,j}$.
We further define a diagonal block matrix $\textbf Q$ as
%\begin{equation}\label{24}
${\textbf{Q}}
\triangleq
\texttt{Bdiag}
(\{[\textbf{Q}_{j}]\}_{j\in \mathcal{V}}
$
%\end{equation}
with  increasing order  on $j$,
and  $\textbf{v}^{\left(\ell\right)}$ and $\textbf{b}$ be the vectors  containing $\textbf{v}_{j }$ and
$\textbf{b}_{j }$, respectively, with the same stacking order as $\textbf{Q}_{j} $.
Following (\ref{v2fm4}), we have
\begin{equation}\label{meanvectorupdate}
\textbf{v}^{\left(\ell\right)} =
 -\textbf{Q} \textbf{v}^{\left(\ell-1\right)} + \textbf{b}.
\end{equation}
For this linear updating equation,
it is well known that, for arbitrary initial value $\textbf{v}^{\left(0\right)}$, $\textbf{v}^{\left(\ell\right)}$ converges
if and only if the spectral radius $\rho\left(\textbf{Q}\right)<1$.
Note that an algorithmically we to check this condition in a distributed manner is provided in \cite{duAsilomar17}.
As convergence of $\textbf{v}^{\left(\ell\right)}$ depends on the convergence of  $\textbf{C}^{\left(\ell\right)}$, we have the following result.
\begin{mytheorem} \label{meanvector}
The vector sequence
$\left\{\textbf{v}^{\left(\ell\right)}\right\}_{l=0,1,\ldots}$ defined by (\ref{meanvectorupdate}) converges to a unique value
for any initial value $\left\{\textbf{v}^{\left(0\right)}\right\}$ and initial covariance matrix $\textbf{C}^{\left(0\right)}\succeq \mathbf 0$ if and only if $\rho\left(\mathbf {Q}\right)<1$.
\end{mytheorem}

According to (\ref{beliefu}), the convergence of
$\bm \mu_i^{(l)}$ depends on
$\big[\textbf C_{f_{i,j}\to i}^{(l)}\big]^{-1}$ and $\textbf v^{(l)}_{f_{i,j}\to i}$.
As Theorem 1 shows that $\big[\textbf C_{f_{i,j}\to i}^{(l)}\big]^{-1}$ is convergence guaranteed with arbitrary positive semidefinite initial value,
the convergence condition of $\bm \mu_i^{(l)}$ is equivalent to the convergence of $\textbf v^{(l)}_{f_{i,j}\to i}$.
Moreover, as shown in \cite{journalversion},  once $\bm \mu_i^{(l)}$ converges, it
converges to
$\hat{\textbf x}_i$.
We therefore conclude that the necessary and sufficient convergence condition of $\bm \mu_i^{(l)}$ to the optimal estimate is
$\rho\left(\mathbf {Q}\right)<1$.

\section{Conclusion}\label{conclusion}
In this paper, we have studied distributed inference using
 Gaussian belief propagation (BP) over networks with two neighboring agents sharing a common observation.
We have analyzed the
 convergence property  of the
Gaussian BP algorithm for this particular model.
We have shown analytically that,
with arbitrary positive
semidefinite matrix  initialization,
the  message information matrix exchanged among agents  converges  at a geometric rate to a unique positive definite  matrix.
Moreover, we have presented  the necessary and sufficient  condition for convergence under which the belief mean
vector  converges to the optimal centralized estimate.

\vfill\pagebreak

\end{document}